\documentclass[sigconf]{acmart}

\def\BibTeX{{\rm B\kern-.05em{\sc i\kern-.025em b}\kern-.08emT\kern-.1667em\lower.7ex\hbox{E}\kern-.125emX}}
    
\setcopyright{none}
\renewcommand\footnotetextcopyrightpermission[1]{}

\usepackage{booktabs} 
\usepackage{graphicx}  
\usepackage{subcaption}
\usepackage{multirow}
\usepackage[ruled,linesnumbered]{algorithm2e}
\newtheorem{assump}{Assumption}
\usepackage{color}
\usepackage{diagbox}
\begin{document}

%
\title{Learning to Advertise for Organic Traffic Maximization\\in E-Commerce Product Feeds}

%

\author{Dagui Chen}
\email{dagui.cdg@alibaba-inc.com}
\affiliation{%
	\institution{Alibaba Group}
}

\author{Junqi Jin}
\email{junqi.jjq@alibaba-inc.com}
\affiliation{%
	\institution{Alibaba Group}
}

\author{Weinan Zhang}
\email{wnzhang@sjtu.edu.cn}
\affiliation{%
	\institution{Shanghai Jiao Tong University}
}

\author{Fei Pan, Lvyin Niu, Chuan Yu}
\email{{pf88537,lvyin.nly,yuchuan.yc}@alibaba-inc.com}
\affiliation{%
	\institution{Alibaba Group}
}

\author{Jun Wang}
\email{jun.wang@cs.ucl.ac.uk}
\affiliation{%
	\institution{University College London}
}

\author{Han Li, Jian Xu, Kun Gai}
\email{{lihan.lh, xiyu.xj, jingshi.gk}@alibaba-inc.com}
\affiliation{%
	\institution{Alibaba Group}
}

%
\renewcommand{\shortauthors}{Dagui and Junqi et al.}
\renewcommand{\shorttitle}{E-Commerce Organic Traffic Maximization}
\settopmatter{printacmref=false}

\begin{abstract}
	Most e-commerce product feeds provide blended results of advertised products and recommended products to consumers.
	The underlying advertising and recommendation platforms share similar if not exactly the same set of candidate products.
	Consumers’ behaviors on the advertised results constitute part of the recommendation model's training data and therefore can influence the recommended results.
	We refer to this process as \emph{Leverage}.
	Considering this mechanism, we propose a novel perspective that advertisers can strategically bid through the advertising platform to optimize their recommended organic traffic.
	By analyzing the real-world data, we first explain the principles of Leverage mechanism, i.e., the dynamic models of Leverage.
	Then we introduce a novel Leverage optimization problem and formulate it with a Markov Decision Process.
	To deal with the sample complexity challenge in model-free reinforcement learning, we propose a novel Hybrid Training Leverage Bidding (HTLB) algorithm which combines the real-world samples and the emulator-generated samples to boost the learning speed and stability.
	Our offline experiments as well as the results from the online deployment demonstrate the superior performance of our approach.
\end{abstract}

%
%


%
%

%
\maketitle

\section{Introduction}\label{sec:intro}


\footnote{This paper has been accepted by CIKM2019}
With the emergence of smartphones, more and more e-commerce companies, such as Amazon, eBay, and Taobao, have adopted the product feeds in their mobile websites and apps.
The products displayed in the feed are usually a mixture of advertised products and recommended products, both of which are selected from similar if not exactly the same set of candidate products~\cite{evans2009online,goldfarb2011online}.
To minimize disruption to consumers' online experience, the two types of products have similar forms of presentation and are blended with a merged layout~\cite{campbell2015good,zhou2016predicting}.

When a consumer requests the e-commerce service, the advertising and recommendation platforms need to determine which advertised products and recommended products should be displayed, respectively.
This process is widely known as \emph{traffic allocation}.
Hereinafter, the traffic from the advertising platform is called ``business traffic'', and the traffic from the recommendation platform is called ``organic traffic''.
As for the recommendation, a score is calculated by a machine learning model for each consumer-product pair.
The top-scored products will be displayed to the consumer aiming to promote user experience related indices, such as trading volume and relevance.
The parameters of the model are trained via recent consumers' behavioral data such as \emph{click}, \emph{buy}.
Compared to the recommendation platform, the advertising platform also computes these historical-data based scores,
but it mainly aims to help advertisers reach targeted consumers and is required to consider the bid prices~\cite{evans2009online,goldfarb2011online} set by advertisers.
Since the bid prices are used along with the scores to determine the advertising ranking,
the advertisers are allowed to design their bidding strategies to compete for business traffic.

\begin{figure}
	\centering
	\includegraphics[width=\linewidth]{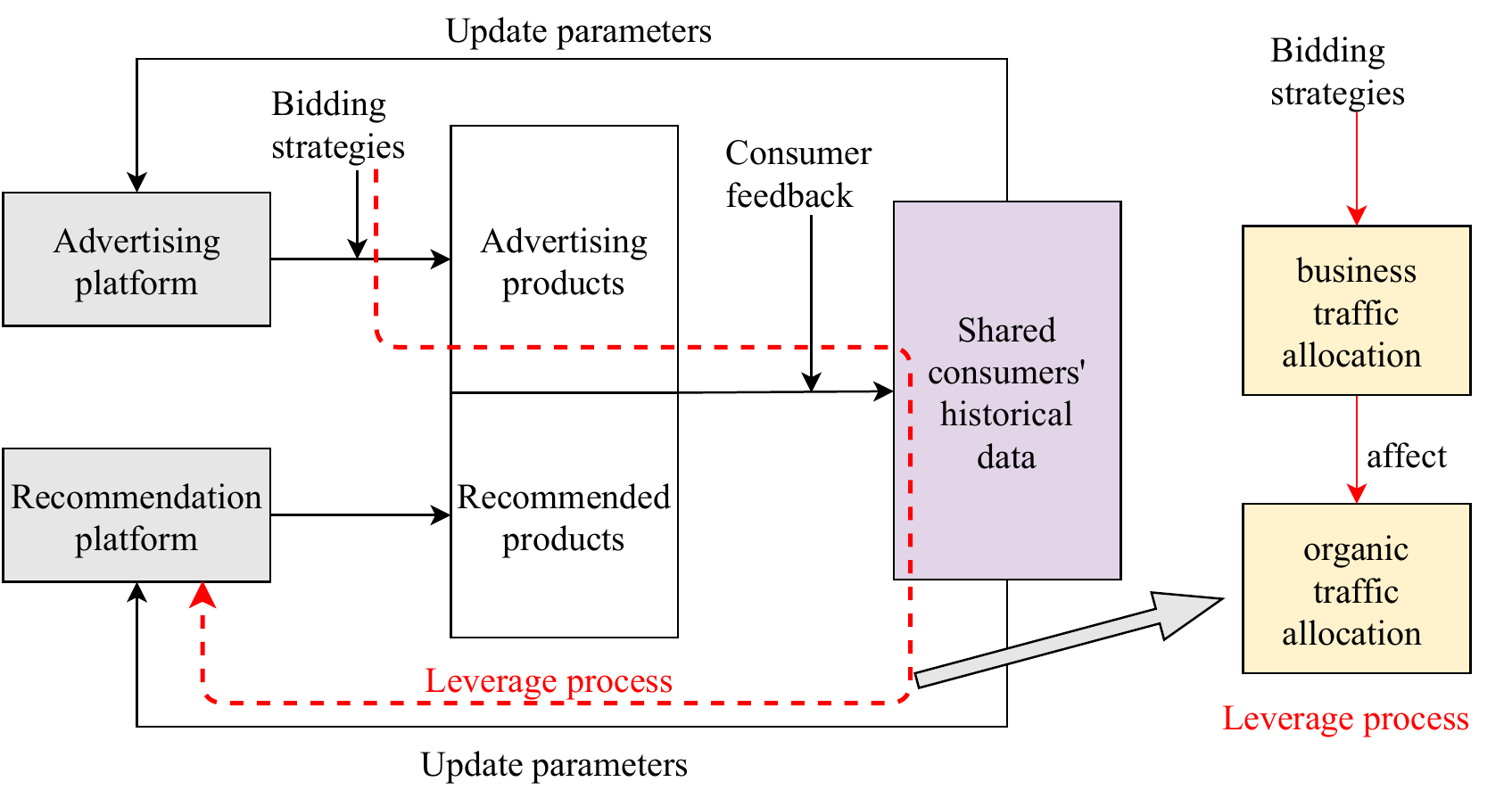}
	\caption{
		Schematic diagram of the Leverage process.
	}
	\label{fig:leverage_process}
\end{figure}

Although these two different platforms operate recommendation and advertisement, respectively, they are highly dependent.
First, they select products from a shared products' set,
so a product can be displayed through either or both platforms.
Second, both platforms use the shared consumers' behavioral histories as data to train their machine learning models.
In one consumer's request round, the results delivered by the advertising platform might affect the record histories of the products in the subsequent rounds.
As such, the changed record data are used to update the parameters of the model in the recommendation platform,
which further influences the recommended scores and finally changes the organic traffic allocation (see Figure~\ref{fig:leverage_process}).
In this paper, we refer to this mechanism, i.e., ``Business traffic allocation can affect organic traffic allocation'', as \emph{Leverage}.

Advertisers always desire more organic traffic because the traffic coming from a platform's voluntary recommendation is usually free.
However, there is no direct way for advertisers to achieve it in the past.
Traditional research papers~\cite{cai2017real,wang2017ladder,Optimizedcostper2017Zhu,perlich2012bid,Optimalrealtime2014Zhang,wu2018budget} on optimal bidding focus on optimizing only the advertising platform's unilateral indices (e.g., page view, click, buy) without considering the effects on other platforms.
In this paper, a novel perspective is proposed that advertisers can bid strategically to compete for not only business traffic but also organic traffic by utilizing the Leverage mechanism.
Apart from the traditional indices in advertising campaigns, we consider a novel optimizing index, i.e., \emph{leveraged traffic},
which represents the expected additional organic traffic that can be obtained through advertising.
With the optimized bidding strategies that maximize the leveraged traffic, advertisers can obtain more potential economic benefits.
The optimization of this index is referred to as \emph{Leverage optimization problem}.

As far as we know, this is the first work to propose the Leverage optimization problem in e-commerce applications.
It requires algorithms to be able to evaluate consumers' request and then adjust advertisers' bid prices to maximize the advertisers' cumulative organic traffic over a period.
However, solving this problem is very challenging, and we list several reasons as well as our proposed solutions as follows.

First, the machine learning model of recommendation platform is trained using several days' historical data.
Thus, any one-shot advertising result's change caused by the bidding algorithm cannot influence organic traffic allocation immediately.
Only several days' sequential bidding decisions can affect the recommendation's model and acquire delayed rewards.
To handle such sequential decision making and delayed reward problem, we use a Markov Decision Process (MDP) as the modeling framework.

Second, from the advertising platform's view, the detailed internal process of the recommendation platform is unknown.
To help understand the principle of Leverage mechanism, we first explain the dynamic process of organic traffic by analyzing real-world observational data and then demonstrate typical Leverage cases.
Moreover, we reveal the characteristics of the Leverage optimization problem and its difference from traditional MDPs.
With the unknown recommendation settings, we formulate the problem via a model-free reinforcement learning (RL) method.

Finally, although model-free RL algorithms~\cite{sutton1998reinforcement,mnih2015human} can solve MDPs with unknown environments, they suffer from high sample complexity and insufficient exploration, especially in real-world applications.
In this paper, we suggest a novel Hybrid Training Leverage Bidding (HTLB) algorithm which takes advantage of the domain properties (see Section~\ref{sec:htlb}) of the Leverage optimization problem.
The proposed algorithm decomposes the agent's state into advertisement-related state and recommendation-related state and uses the real-world and emulator-generated hybrid-sample to learn the bidding strategies.
Our HTLB method shows an obvious boost on learning speed and stability, which is demonstrated by our extensive experiments on the real-world data.

In sum, our contributions are:
(i) We first reveal the Leverage mechanism and then propose a novel Leverage optimization problem in e-commerce product feeds.
(ii) We first explain the principle of Leverage mechanism through the analysis of the real-world data.
(iii) We formulate the Leverage optimization problem as a model-free MDP and introduce a novel HTLB algorithm which can reduce the sample complexity significantly.
(iv) We validate the efficacy and efficiency of our approach with extensive offline and online experiments. The algorithm to optimize Leverage has been now deployed in Taobao's display advertising system.

\section{Related Works}\label{sec:rw}
\subsection{Relationship between Rec. and Ad.}

In an online service with dependent platforms, the advertisement and the recommendation interact in different aspects.
\citet{danescu2010competing} studied the competitive and synergistic interplay between advertising and organic results on the search engine results page.
\citet{Analyzingtherelationship2010Yang} suggested that the click-through rate on organic search listings had a positive dependence with the click-through rate on paid listings, and vice versa.
\citet{Doorganicresults2015Agarwal} found that an increase in organic competition led to a decrease in click performance, but helped the conversion performance of sponsored ads.
By utilizing a hierarchical Bayesian model, \citet{agarwal2011sponsored} showed that organic links in sponsored search might hurt performances for immediate transactions but increase the brand awareness for the advertisers.
As for advertisers' bidding mechanism, \citet{Effectsofthe2012Xu} analyzed how the presence of organic listing as a competing information source affected advertisers' sponsored bidding and the equilibrium outcomes in search advertising.

Nevertheless, all above works only considered the impact of organic and business traffic on consumers or advertisers.
None of them dived into the interaction effects of different platforms on final performance.
This work considers both the interplay of advertising and recommendation platforms and the consumers' feedback.
Moreover, our work not only reveals the dependent relationship but also studies how to utilize this relationship to optimize the recommendation platform's performance.

\subsection{RL methods for Bidding Strategies}
Since the bidding strategies optimization in online advertising could be modeled as a sequential decision problem,
several works utilized RL methods to solve it.
\citet{cai2018reinforcement} formulated the impression allocation problem as an MDP and solved it by an actor-critic policy gradient algorithm based on DDPG.
\citet{cai2017real} formulated a Markov Decision Process framework to learn sequential allocation of campaign budgets.
\citet{hu2018reinforcement} formulated the multi-step ranking problem as a search session MDP and proposed a deep policy gradient algorithm to solve the problem of high reward variance and unbalanced reward distribution.
\citet{wang2017ladder} utilized deep Q network (DQN) to optimize the bidding strategy in DSP.
\citet{RealTimeBidding2018Jin} formulated bidding optimization with multi-agent reinforcement learning to balance the trade-off between the competition and cooperation among advertisers.

However, the above bidding algorithms only focused on the advertising platform's performance, and their model-free RL methods suffered from the high sample complexity challenge.
Compared to these traditional optimization views, this paper proposes a novel optimizing index, i.e., leveraged traffic, and proposes formulation for the Leverage optimization problem.
Furthermore, to alleviate the high sample complexity, we introduce the HTLB algorithm to augment training samples, which achieves better convergence and stability.

\section{Evaluation Platform}\label{sec:set}

\begin{figure}
	\centering
	\includegraphics[width=\linewidth]{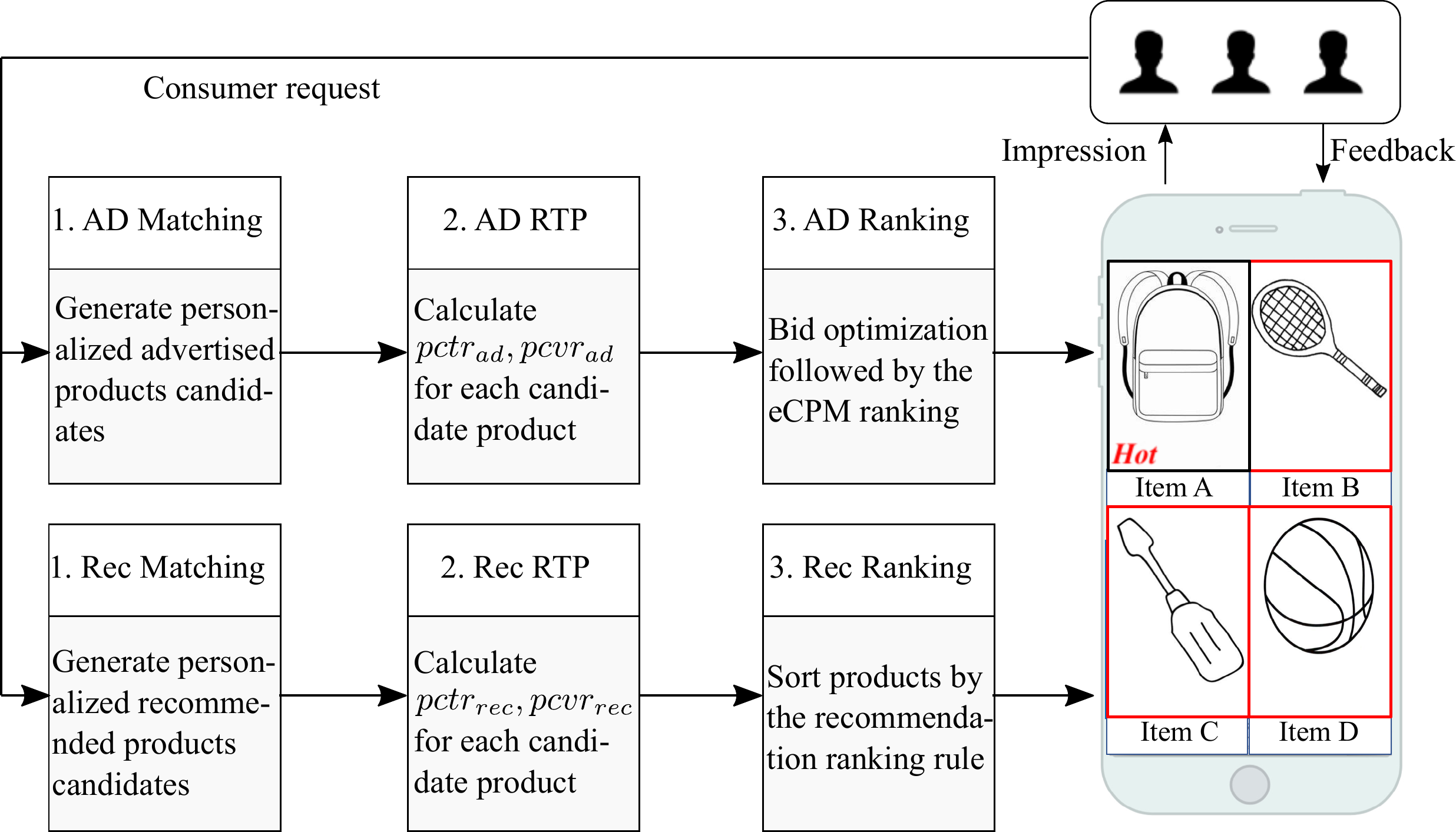}
	\caption{
		An overview of \emph{Guess What You Like} of Taobao App.
		The displayed advertised products are tagged with ``Hot'' and surrounded by the recommended results.
	}
	\label{fig:taobao_system}
\end{figure}

Without loss of generality, we choose one of the largest e-commerce platform in China, Taobao, as the testbed to analyze the Leverage mechanism and evaluate our algorithm.
It is worth mentioning that our analysis and proposed bidding algorithm can easily generalize to other content feeds with blended organic traffic and business traffic.
\emph{Guess What You Like} is a typical example of e-commerce product feeds located at the homepage of Taobao Mobile App, where three display advertising slots and hundreds of recommended slots are well organized in a structured way in response to the consumers' requests.
The recommendation and advertising platforms both independently contain three similar modules: matching, real-time prediction (RTP) and ranking~(see Figure~\ref{fig:taobao_system}).

\textbf{Advertising}. After receiving the consumer request, the advertising matching module first recalls hundreds of personalized advertised product candidates by analyzing the relevance between the products and the consumer's preference and incorporating the advertisers' willingness of competing for the impression.
Second, the follow-up advertising RTP engine calculates the predicted click-through rate ($pctr$) and the predicted conversion rate ($pcvr$) for each eligible advertised product.
After that, the advertising ranking module considers the bid prices set by the advertisers and then ranks the candidate advertised products by the descending order of $pctr \times bid$, which is called \emph{effective Cost Per Mille} (eCPM) sorting mechanism.
Finally, the top-ranked products are displayed.
Because the advertising platform has more complete consumers' data and the advertising campaigns' information, for better bidding strategies,
the advertising platform is authorized to adjust the bid with a scale factor $\alpha$ to generate $adjust\_bid = bid \times (1+ \alpha)$,
where $\alpha \in [-range, range]$ and $range$ is the adjust ratio bound.
Thus, the eCPM ranking rule becomes $pctr \times adjust\_bid$.
This scale factor therefore is used to optimize advertisers' bidding strategies in some researches~\cite{Optimizedcostper2017Zhu,RealTimeBidding2018Jin}.

\textbf{Recommendation}. The recommendation platform has a similar process to the advertising platform.
The main differences are:
(i) The recommendation platform mainly aims to enhance the consumer experience such as optimizing the total trading volume and relevance and thus has a different ranking rule, e.g., maximizing trading volume.
(ii) To ensure the consumer experience, the number of advertised products displayed in one consumer request is strictly limited.
Therefore, the recommendation platform contains more recalled candidates in the matching stage and more display slots.
(iii) More importantly, there is no advertiser participation in the recommendation system.
Thus, the recommendation platform does not take the advertisers' demands into account,
which also implies that the advertisers have no direct way to compete for more organic traffic.

\section{Leverage Mechanism}\label{sec:data-an}

\subsection{Different Leverage Phenomena}
To study the Leverage mechanism, we observe the products advertised on \emph{Guess What You Like} and compare their average organic traffic per day in three stages: before advertising, during advertising, and after advertising,
denoted as \emph{be-ad}, \emph{wh-ad}, and \emph{af-ad}, respectively.
Considering that the organic traffic can be influenced by other factors, we choose 1,000 products whose organic traffic remain stable (i.e., traffic variance less than a threshold) for three days before advertising.

We analyze the organic traffic trend of products from three aspects.
First, by comparing \emph{wh-ad} and \emph{be-ad}, we explore the impact of advertising campaigns on organic traffic.
Next, we observe whether their organic traffic will decline after the advertiser stops advertising.
Finally, we pick out the products whose organic traffic has dropped in \emph{af-ad} by more than 10\% and then look into whether their performances return to the level before advertising.
In each comparison, we divide the relative increments of organic traffic into different intervals.
The number of products located in each interval is counted and recorded in Table~\ref{tab:lever_pheno}.
\begin{table}[tb]
	\caption{
		The number of products with different Leverage phenomena.
		Here, 52 and 250 are bolded to indicate that a total of 302~($52 + 250$) items have a significant \emph{af-ad} organic traffic drop.
		The statistics in the last column~(i.e., \emph{af-ad} vs. \emph{be-ad}) are only for these items, i.e., $131 + 171 = 52 + 250$.
	}\label{tab:lever_pheno}
\small{
\begin{tabular}{@{}cccc@{}}
\toprule
\% improve&\emph{wh-ad} vs. \emph{be-ad} &\emph{af-ad} vs. \emph{wh-ad}&\emph{af-ad} vs. \emph{be-ad} \\
\midrule
$<-50\%$&11&\textbf{52}&\multirow{3}{*}{131}\\
$(-50\%,-10\%]$&192&\textbf{250}&\\
$(-10\%,0]$&73&97&\\\midrule
$(0,10\%]$&53&87&\multirow{3}{*}{171}\\
$(10\%,50\%]$&209&197&\\
$>50\%$&462&317&\\\bottomrule
\end{tabular}}
\end{table}


As shown in Table~\ref{tab:lever_pheno}, the 1000 products here show different Leverage phenomena.
Compared to \emph{be-ad}, most products' (i.e., $(53+209+462)/1000= 72.4\%$) organic traffic has increased during \emph{wh-ad},
and 462 of them achieve more than 50\% improvement,
which verifies that the delivery of advertisement does affect the organic traffic allocation and most of the impact is positive.
It is worth noting that the improvement does not vanish after the removal of the advertisement and even 601~(i.e., $87 + 197 + 317$) products gained new organic traffic growth,
which means that advertising has the potential to assist these products in escaping the ``cold start''stage in the recommendation platform.
As shown in the last column in Table~\ref{tab:lever_pheno}, we find that although 302 (i.e., $52+250$ or $131+171$) items have a significant \emph{af-ad} organic traffic drop compared to \emph{wh-ad}, there are still 171 items performing better than \emph{be-ad},
which implies that advertising has a long-term impact on some products' organic traffic.

\begin{figure*}[htpb]
	\centering
	\includegraphics[width=\linewidth]{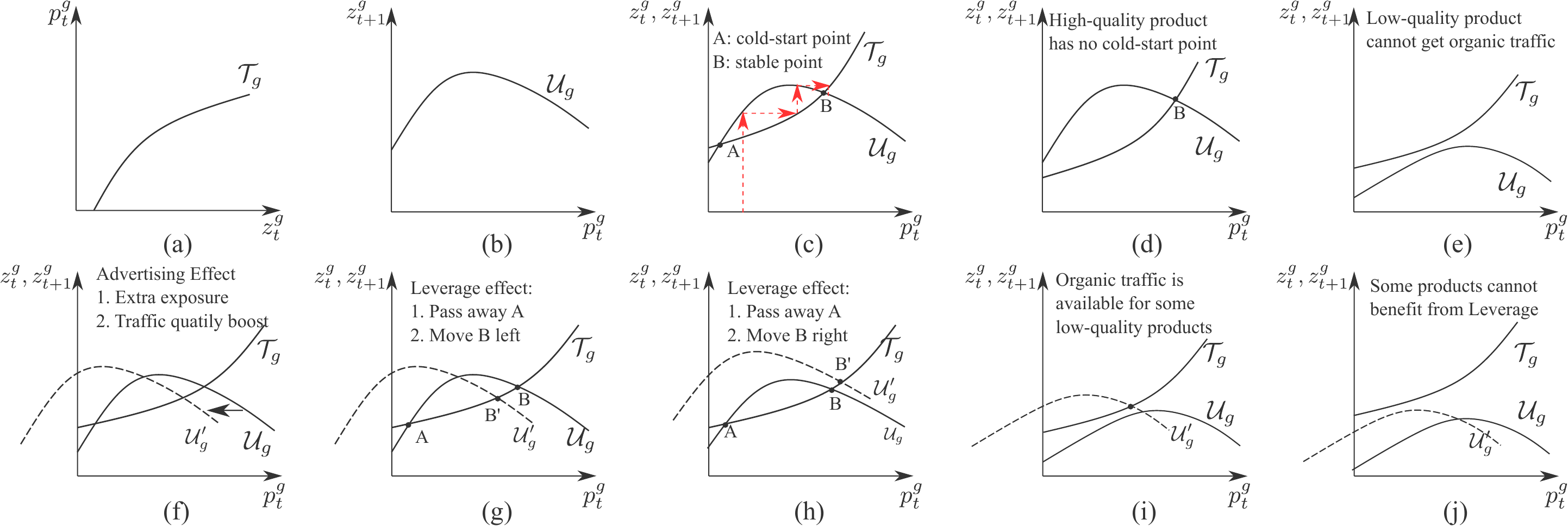}
	\caption{
		The schematic diagram of the Leverage principle.
		$z^g$ is the average recommended score for product $g$ and $p^g$ is its organic traffic.
		(a) The curve of traffic-win function $\mathcal{T}_g$.
		(b) One typical exposure-effect function $\mathcal{U}_g$. 
		(c-e) The products with different attributes have different dynamic processes under the “stable condition”. 
		(f) The advertising can change the dynamics by moving curve $\mathcal{U}_g$.
		(g-j) Different typical Leverage cases can be demonstrated.
		The details of all subfigures refer to the text.
	}\label{fig:lever_principle}
\end{figure*}

\subsection{Leverage Principle and Dynamic Process}
The mechanism that supports the Leverage phenomenon is that the products' exposure to consumers~(i.e., the traffic) can influence their properties.
To explain the Leverage phenomenon, this subsection will model the relationship between the amount of exposure and the average recommended score.
After that, we will reveal the Leverage principle under the hood, i.e., the dynamic model of Leverage which can explain different Leverage phenomena mentioned above.


First, we describe the dynamics of the recommendation platform under an ideal stable condition, denoted as \emph{stable condition}
which means that \emph{the recommendation platform's strategy and the consumers who request services remain fixed, and there is no intervention from the advertising platform}.
Then we explain how advertising changes the \emph{stable condition} and causes different Leverage phenomena.
It is worth mentioning that all the following text in this subsection is the explanation for Figure~\ref{fig:lever_principle}.

To simplify our description, for a specific product $g$, we denote its amount of recommended exposure (i.e., organic traffic) in $t$-th time window as $p_t^g$
and the corresponding average recommended score calculated by the recommendation's RTP engine as $z_t^g$.
Next, we describe the relationship between $z^g$ and $p^g$ through two assumptions.
\begin{assump}\label{assump:1}
	For a specified product $g$, under the \emph{stable condition}, the amount of exposure $p^{g}_t$ is nonlinearly and monotonically increasing with respect to average recommended score $z_t^g$, which is denoted by $p^g_t \triangleq \mathcal{T}_g (z^g_t)$, where $\mathcal{T}_g$ is called ``traffic-win" function.
\end{assump}

An example of the traffic-win curve is shown as Figure~\ref{fig:lever_principle}a.
This curve is somewhat reasonable due to the following facts.
(i) If the product is evaluated with a higher score by the recommendation's RTP, it can obtain more expected organic traffic.
(ii) The increasing trend is similar to the advertisement's win-rate curve with respect to the bid price~\cite{Optimalrealtime2014Zhang} in the advertising platform
because the bid price plays the same role (i.e., multiplicative factor) in the advertising ranking as the recommended score $z^g_t$ in the recommendation's ranking.
(iii) Due to the truncation effect of the ranking rule, the product is not able to obtain organic traffic unless $z^g_t$ is greater than a certain threshold.
The Assumption~\ref{assump:1} describes the impact of the average recommended score on organic traffic allocation,
while the following assumption states the effect of the amount of exposure on the recommended score.
\begin{assump}\label{assump:2}
	For a specified product $g$, under the \emph{stable condition}, the next average recommended score $z_{t+1}^g$ can be influenced by its amount of exposure $p_t^g$.
	Moreover, the relationship satisfies the Markov property.
	\begin{equation}
		z_{t+1}^g \triangleq \mathcal{U}_g(p_1^g, p_2^g, \cdots, p_t^g)=\mathcal{U}_g(p_t^g)
		\label{eq:pv_pctr}
	\end{equation}
	where the $\mathcal{U}_g$ is called ``exposure-effect'' function.
\end{assump}

$\mathcal{U}_g$ represents the process of consumers' feedback behaviors.
These behaviors data update the RTP model's parameters,
which further affects the next evaluation of the RTP model.
To study the specific form of $\mathcal{U}_g$, we have conducted preliminary experiments on the real-world data to observe different products' exposure effects.
Based on our preliminary experiments, we find out that the forms of $\mathcal{U}_g$ vary among the products.
One typical $\mathcal{U}_g$ shows a trend of rising and then decreasing, as shown in Figure~\ref{fig:lever_principle}b.
The reason of the rise is that a certain amount of exposure would attract the behavior of the consumers.
However, the consumers' behavior will be diluted by too much organic exposure in the case of stable market,
which leads to the decreasing trend of the second half of the curve.

Based on the above two assumptions, the sequence $\{p_0^g$, $z_1^g = \mathcal{U}_g (p_0^g)$, $p_1^g = \mathcal{T}_g(z_1^g)$, $z_2^g =\mathcal{U}_g (p_1^g)$, $p_2^g = \mathcal{T}_g (z_2^g)$, $\ldots\}$ constitutes a Markov chain which describes the dynamic relationship between the organic traffic and the average recommended score.
Figure~\ref{fig:lever_principle}c reorganizes the axes and merges $z_t^g$ and $z_{t+1}^g$ so that $\mathcal{T}_g$ and $\mathcal{U}_g$ can be drawn in the same coordinates.
The dashed arrow depicts the dynamic sequence of $p^g$ and $z^g$, and the sequence generally converges to the stable point $B$ whose $p^g_t$-axis coordinate indicates the organic traffic that the product $g$ can obtain under the \emph{stable condition}.
However, if the initial organic traffic $p_0^g$ is on the left side of another intersection point $A$, the product $g$ will not obtain organic traffic, so we call it ``cold start'' point.
Besides, different products have different dynamic processes.
For example, the high-quality products (i.e., higher average recommended score) without any cold start point can always obtain organic traffic (see Figure~\ref{fig:lever_principle}d),
while for low-quality products (see Figure~\ref{fig:lever_principle}e), it is generally difficult to obtain organic traffic.

When the advertiser begins to advertise, the \emph{stable condition} is broken.
The role of advertising is divided into two parts:
(i) The extra business traffic gives an offset on the exposure (i.e., more exposure opportunities), so it requires less organic traffic to obtain the same consumers' feedback, which will move the curve $\mathcal{U}_g$ leftwards.
(ii) Considering that the average properties of business traffic and organic traffic are different, the consumers' feedback on the two are also different.
In most cases, the consumers' feedback on the business traffic is better than the organic traffic, because the advertisers can design good advertising creatives to attract consumers to click and the advertising platform provides extra premium display slots~(e.g., eye-catching location), which will move the curve $\mathcal{U}_g$ upwards.
For the two reasons, the curve $\mathcal{U}_g$ mostly moves towards the upper-left direction when the product is advertised, as shown in Figure~\ref{fig:lever_principle}f.

Based on the previous analysis, different typical Leverage phenomena can be demonstrated.
As shown in Figure~\ref{fig:lever_principle}g and Figure~\ref{fig:lever_principle}h, on the one hand, Leverage can help the products escape the cold start stage in recommendation platform.
On the other hand, the organic traffic at the stable point $B$ will be changed (increase or decrease).
Even for a low-quality product, a certain amount of organic traffic is still available through advertising, as shown in Figure~\ref{fig:lever_principle}i,
but the organic traffic will disappear after the advertisement is removed.
Finally, some products cannot benefit from Leverage, as shown in Figure~\ref{fig:lever_principle}j.

It is worth mentioning that in a few cases, the consumers' interaction on business traffic is lower than on organic traffic.
As a result, the advertising will move the curve $\mathcal{U}_g$ towards the bottom-left direction.
The typical Leverage phenomena can also be verified via similar analysis.
However, in these cases, Leverage is more likely to have a negative effect, where the organic traffic decreases because of advertising.


\section{Leverage Optimization}\label{sec:mdp}
In addition to revealing the Leverage mechanism, another important contribution of this paper is that we first consider utilizing this dependent mechanism to optimize the leveraged traffic from the view of advertising platform, which is denoted as \emph{Leverage optimization}.
As stated in Section~\ref{sec:intro}, it is a sequential decision problem with delayed reward, and we model it with an MDP framework.



\subsection{Optimization Problem as a PKMDP}
A vital feature of the Leverage optimization problem is that it involves two platforms.
Although the two platforms have similar processes for handling consumer requests~(see Figure~\ref{fig:taobao_system}), they differ significantly in terms of bidding strategy optimization, so they have different state transition models.
Considering this, we formulate the Leverage optimization problem as a Partially Known model MDP~(PKMDP) which is defined as a tuple $\mathcal{M} = (\mathcal{S}, \mathcal{A}, \mathcal{P}^o, \mathcal{P}^x, \mathcal{R})$.

$\mathcal{S}$ is the set of states. Each state $s\in \mathcal{S}$ can be decomposed into two parts, i.e., $s\triangleq[\mathbf{o}, \mathbf{x}]$,
where $\mathbf{o}=[o^1, o^2, \ldots, o^K]$ indicates the advertisement-related state
and $\mathbf{x} =[x^1, x^2, \ldots, x^K]$ indicates the recommendation-related state.
$K$ is the number of targeted products.
For a specific product $g$, the advertisement-related state represents the feature recorded in the advertising platform, such as $click_{ad}$.
Correspondingly, the number of clicks that the product receives as a recommended product is included in the recommendation-related state.

$\mathcal{A}$ is the set of actions and each action $a \in \mathcal{A}$ denotes a bidding decision for all targeted products, $a=[\alpha^1, \alpha^2, \cdots, \alpha^K]$.
In our setting, the action of a specific product is the bid adjust ratio $\alpha$ for the fixed bid price (see Section~\ref{sec:set}).

$\mathcal{P}^\mathbf{o} (\mathbf{o}'|[\mathbf{o}, \mathbf{x}], a)$ indicate the probability that the advertisement-related state transitions to $\mathbf{o}'$ conditioned on the state $[\mathbf{o}, \mathbf{x}]$ and the bidding action $a$, while $\mathcal{P}^\mathbf{x} (\mathbf{x}'|[\mathbf{o}, \mathbf{x}], a)$ is for the recommendation-related state.
As the advertising platform, we can build an advertising emulator which replicates the sorting mechanism in the online advertising system.
Therefore, we can obtain the successor advertisement-related state $\mathbf{o}'$ under different bidding strategies through this emulator~(i.e., we can simulate the transition function $\mathcal{P}^\mathbf{o}$).
However, because the detailed internal process of recommendation platform is not clear, the transition function $\mathcal{P}^\mathbf{x}$ is unknown.


$\mathcal{R}$ is the reward function: $\mathcal{R}(s, a)=\sum_{k=1}^K \eta_k r_k (s, a)$, where $r_k (s, a)$ is the increment of leveraged traffic for $k$-th product,
and $\eta_k$ measures the importance of this product.
We can achieve different optimization objectives by choosing the different forms of $\eta$.
For example, we can maximize the total increment of organic traffic when $\eta_k = 1, \forall k$
and maximize the number of products with increased organic traffic when $\eta_k = \frac{1}{|r_k(s,a)|}, \forall k$.

Here we consider a deterministic policy $\pi$, i.e., a mapping $\mathcal{S} \mapsto \mathcal{A}$.
The goal of RL is to find an optimal policy $\pi^*$ to maximize the expected episodic return:
\begin{equation}
	\pi^* = \arg \max_\pi \mathbb{E} \left[ \sum_{t=0}^{H-1} \gamma^t R(s_t, a_t) \big| \pi\right]
	\label{eq:objective}
\end{equation}
where $H$ is the length of episode, and $\gamma \in [0, 1]$ is the discount factor.
\begin{figure}[tb]
  \centering
  \includegraphics[width=\linewidth]{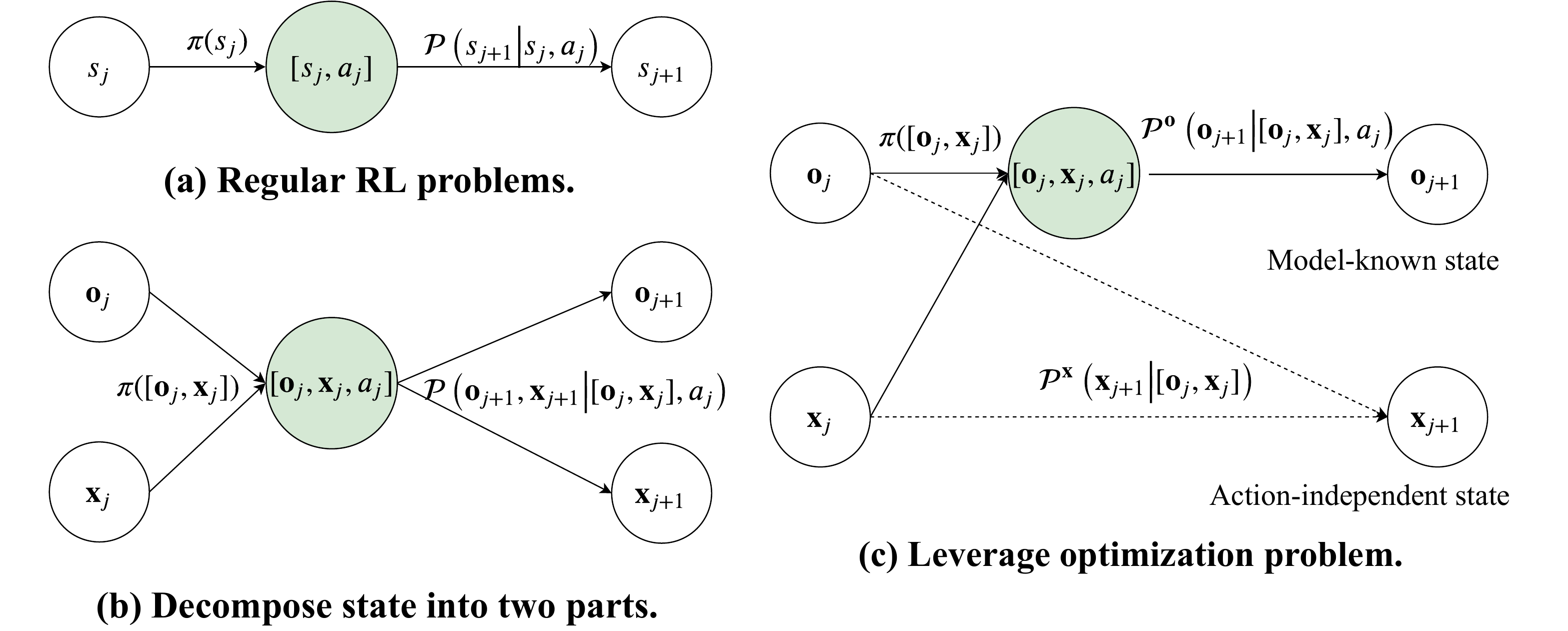}
  \caption{The difference between Leverage optimization problem and the regular RL problems.}\label{fig:leverage_backup}
\end{figure}

For the conventional RL problem, the agent takes action $a$ in state $s$ through the policy $\pi$, and then the agent transitions to a new state $s'$ based on the transition model $\mathcal{P}$~(see Figure~\ref{fig:leverage_backup}a).
As a PKMDP, the Leverage optimization problem follows some unique characteristics.
(i) Different from the regular RL problem, the agent state in our setting can be decomposed into two parts $[\mathbf{o}, \mathbf{x}]$.
The schematic diagram of the decomposition is shown in Figure~\ref{fig:leverage_backup}b.
(ii) Given $[\mathbf{o}_{t}, \mathbf{x}_t]$, the next recommendation-related state $\mathbf{x}_{t+1}$ is independent of action $a_t$,
because our action, i.e., the bid adjust ratio, can only be directly applied to the advertising platform.
The organic traffic allocation is affected only when the prediction model's parameters are updated based on the consumers' behavior on delivered products.
It requires at least two time steps to influence the recommendation-related state through the bidding strategy (i.e., $a_t \rightarrow \mathbf{o}_{t+1} \rightarrow \mathbf{x}_{t+2}$), as shown in Figure~\ref{fig:leverage_backup}c.
Hence, we call $\mathbf{x}$ ``action-independent'' state.
(iii) Different from the recommendation-related state with unknown transition model, the successor advertisement-related state can be simulated.
Hence, we refer to the advertisement-related state as ``model-known'' state.

\subsection{HTLB Training Method as the Solution}\label{sec:htlb}



The model-free RL is often utilized to solve the MDPs with unknown environment model.
Especially, since the action (i.e., bid adjustment ratio) space is continuous in our formulation, we choose DDPG as our base algorithm.
DDPG~\cite{lillicrap2015continuous} is an off-policy actor-critic algorithm to learn a deterministic policy by extending DQN~\cite{mnih2015human} and DPG~\cite{silver2014deterministic}.
This algorithm consists of two components: an actor which parameterizes a deterministic policy $\pi$, and a critic which approximates the action-value $Q$ function.
The DDPG algorithm uses two techniques from DQN, i.e., experience replay (ER) and target networks, to address the learning instability problem.
ER memory stores the transition $\langle s_t, a_t, r_{t+1}, s_{t+1}\rangle$ into a buffer and then a mini-batch is sampled randomly from ER for learning,
which eliminates the temporal correlation and reduces the update variance.
The target networks, $\pi'$ and $Q'$, which are used to update the value of $\pi$ and $Q$, are constrained to change slowly, thereby improving the stability of learning.
Following~\cite{lillicrap2015continuous} and considering characteristics of the Leverage optimization problem, our critic and actor update rules can be written as
\begin{align}
	y_i &= r_i + \gamma Q'\big([\mathbf{o}_{i+1}, \mathbf{x}_{i+1}],\pi'([\mathbf{o}_{i+1}, \mathbf{x}_{i+1}]) \big) \label{eq:target_q} \\
	L(\theta^Q) &= \frac{1}{N} \sum_i  \left( y_i - Q([\mathbf{o}_i, \mathbf{x}_i], a_i) \right)^2 \label{eq:update_q} \\
	\nabla_{\theta^\pi} J(\theta^\pi)  &\approx \frac{1}{N} \sum_i \nabla_{\theta^\pi} \pi([\mathbf{o}_i, \mathbf{x}_i]) \nabla_a Q\big([\mathbf{o}_i, \mathbf{x}_i], \pi([\mathbf{o}_i, \mathbf{x}_i])\big)\label{eq:update_pi}
\end{align}
where $[\cdot, \cdot]$ is the concatenation operator.

\begin{figure}[tb]
	\centering
	\includegraphics[width=\linewidth]{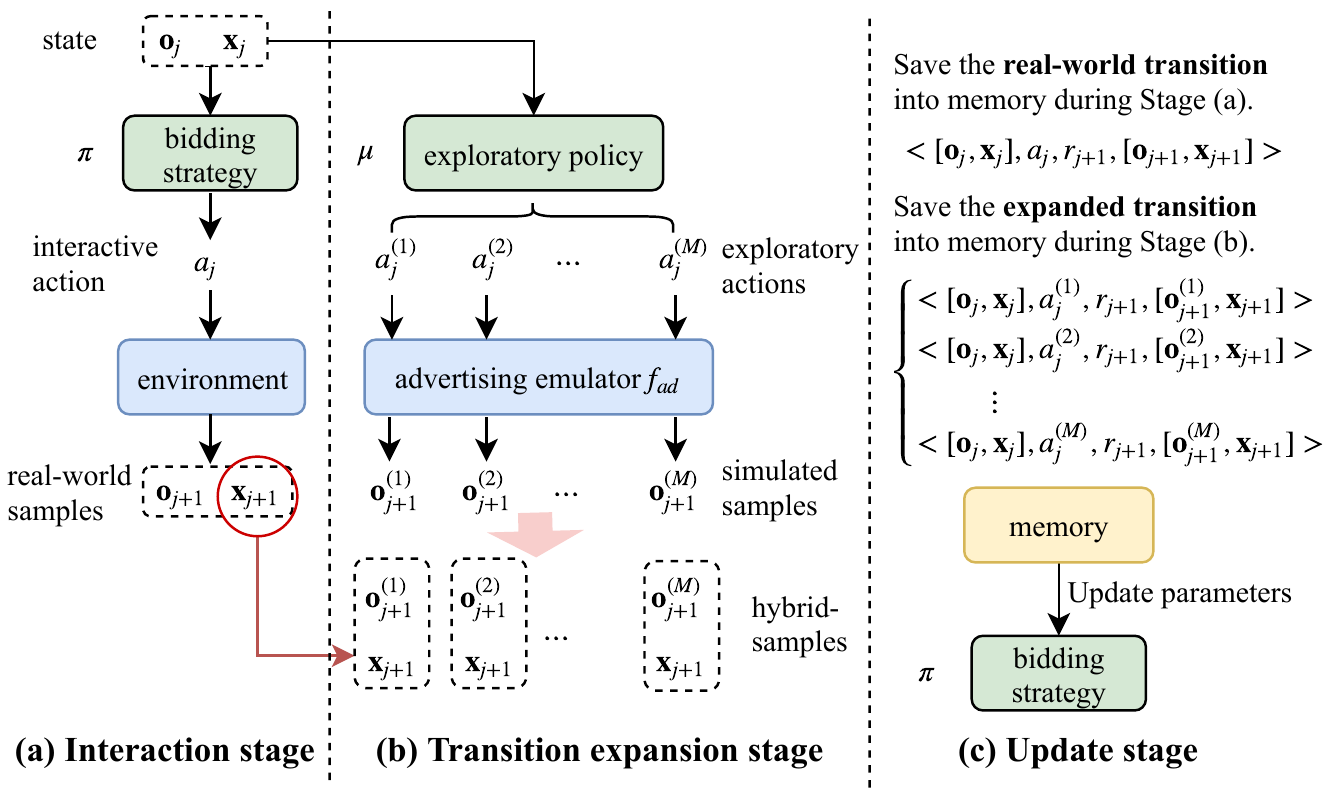}
	\caption{
		The schematic diagram of HTLB.
	}\label{fig:htlb-diagram}
\end{figure}

In general, the model-free RL suffers from high sample complexity and insufficient exploration.
Moreover, in the real-world system, these challenges and the large numbers of samples during the trial-and-error phase may cause much economic and time cost.
In this paper, considering the Leverage optimization as a PKMDP whose states can be decomposed into ``action-independent'' state and ``model-known'' state,
we propose an emulator and real-world hybrid-sample training method which is named Hybrid Training Leverage Bidding to reduce sample complexity effectively.
The method consists of three stages~(see Figure~\ref{fig:htlb-diagram}):
\begin{enumerate}
	\item \textbf{Interaction stage}. Execute target policy $\pi$, observe the successor state $[\mathbf{o}', \mathbf{x}']$ and reward $r$, and save the transition into the memory.
	\item \textbf{Transition expansion stage}. For each transition in the memory $\langle[\mathbf{o}_j, \mathbf{x}_j], a_j, r_{j+1}, [\mathbf{o}_{j+1}, \mathbf{x}_{j+1}]\rangle$,
		sample $M$ actions $a_j^{(1)}, a_j^{(2)}, \ldots, a_j^{(M)}$ from the exploratory policy $\mu$ and expand the memory via the advertising emulator $f_{ad}$ as follows:
		\begin{equation}
			\begin{cases}
				\langle[\mathbf{o}_j, \mathbf{x}_j], a_j^{(1)}, r_{j+1}, [f_{ad}([\mathbf{o}_j, \mathbf{x}_j], a_j^{(1)}), \mathbf{x}_{j+1}]\rangle \\
				\langle[\mathbf{o}_j, \mathbf{x}_j], a_j^{(2)}, r_{j+1}, [f_{ad}([\mathbf{o}_j, \mathbf{x}_j], a_j^{(2)}), \mathbf{x}_{j+1}]\rangle \\
				\qquad \vdots \\
				\langle[\mathbf{o}_j, \mathbf{x}_j], a_j^{(M)}, r_{j+1}, [f_{ad}([\mathbf{o}_j, \mathbf{x}_j], a_j^{(M)}), \mathbf{x}_{j+1}]\rangle,
			\end{cases}
		\end{equation}
where $M$ represents the times of expansion and $f_{ad}$ indicates simulating the successor ``model-known'' state via the advertising emulator.
	\item \textbf{Update stage}. Use the expanded memory to update the parameters of the agent.
\end{enumerate}

\begin{algorithm}[t]
\caption{HTLB-DDPG}\label{alg:HTLB-DDPG}
\small
Initialize critic $Q$ with $\theta^Q$ and actor $\pi$ with $\theta^\pi$\;
Initialize the target network $Q'$, $\pi'$ with weights $\theta^{Q'} \leftarrow \theta^{Q}$, $\theta^{\pi'} \leftarrow \theta^{\pi}$\;
Initialize ER memory $D$, and advertising emulator $f_{ad}$ \;
\For{episode = 1 to E}{
	Initialize a random process $\mathcal{N}$ for exploration\;
	Receive initial state $[\mathbf{o}, \mathbf{x}]$ \;
    \For{t=1 to T}{
		Compute action (bid adjust ratio) via $a_t = \pi([\mathbf{o}_t, \mathbf{x}_t]|\theta^\pi)$ and execute $a_t$ \;
		Observe the next state $[\mathbf{o}_{t+1}, \mathbf{x}_{t+1}]$ and reward $r_{t+1}$, and then
		store the transition $\langle [\mathbf{o}_t, \mathbf{x}_t], a_t, r_{t+1}, [\mathbf{o}_{t+1}, \mathbf{x}_{t+1}]\rangle$ into $D$\;
		\For{m=1 to M}{
			Select the exploratory action $a_t^{(m)} = a_t + \mathcal{N}_t^{(m)}$~(i.e., exploratory policy $\mu$) \;
			Simulate $a_t^{(m)}$ in $f_{ad}$ and get $\mathbf{o}_{t+1}^{(m)}$ \;
			Store the expanded transition $\langle [\mathbf{o}_t, \mathbf{x}_t], a_t^{(m)}, r_{t+1}, [\mathbf{o}_{t+1}^{(m)}, \mathbf{x}_{t+1}]\rangle$ into $D$\;
		}
		Sample a random minibatch of $N$ transitions $\langle [\mathbf{o}_i, \mathbf{x}_i], a_i, r_{i+1}, [\mathbf{o}_{i+1}, \mathbf{x}_{i+1}]\rangle$ \;
		Update $\theta^Q$ by minimizing loss with Eqs. (\ref{eq:target_q}) (\ref{eq:update_q})\;
		Update $\theta^\pi$ with Eq.~(\ref{eq:update_pi}) \;
		Update target network: $\theta' \leftarrow \tau \theta + (1- \tau) \theta'$ \;
    }
}
\end{algorithm}
Our HTLB method is essentially the extension of the Dyna~\cite{sutton1998reinforcement} framework.
By effectively integrating model-free reinforcement learning and model-based planning, Dyna enables the agent to learn value function~(or policy) from both real and simulated experience.
However, the application of Dyna relies on a precise emulator, which requires the transition model to be known or easy to learn.
In the Leverage optimization problem, although we only know the transition model of partial state (i.e., advertisement-related state), the other state (i.e., recommendation-related state) is independent of the action.
Hence, we can combine the simulated advertisement-related sample $f_{ad}([\mathbf{o}_j, \mathbf{x}_j], a_j^{(m)})$ with the real recommendation-related sample $\mathbf{x}_{j+1}$,
to produce large numbers of hybrid-samples $s_{j+1}^{(m)} = [f_{ad}([\mathbf{o}_j, \mathbf{x}_j], a_j^{(m)}), \mathbf{x}_{j+1}]$.


The critical step of our proposed HTLB method is the \emph{transition expansion} stage.
At this stage, we can conduct experiments with different exploratory actions without interacting with the real environment.
For the DDPG algorithm, the transition expansion stage is executed in the form of expanding the experience replay, i.e., the hybrid-samples are also stored into ER for training.
The complete HTLB-DDPG is shown in Algorithm~\ref{alg:HTLB-DDPG}.

Intuitively, our proposed algorithm has two advantages over DDPG.
On the one hand, since the number of trainable transition can be expanded by the advertising emulator,
the real sample complexity is significantly reduced, which leads to faster convergence and lower variance.
On the other hand, the trial-and-error exploration (i.e., $a_t^{(m)}$) is restricted inside the advertising emulator, which can avoid much economic and time cost.

\section{Experiments}\label{sec:exp}
\subsection{Experimental Setup}
\textbf{Time-step and episode length settings}.
In the online advertising system, the ideal way is to evaluate each request to optimize the total leveraged traffic.
However, this arrangement is computationally expensive, and in fact, the exposure of a specific product on each request is sparse.
Besides, the recommendation's RTP module updates its parameter, not after each impression, but based on the overall performance of the product within a certain time window.
Thus, in our settings, the time step $t$ is defined as a time window with equal length (i.e., one day).
The agent determines the action based on the aggregated state on the previous time window and applies it to the next time window.
Besides, one episode contains seven time steps which are the number of state transition.

\textbf{Offline evaluation flow}.
Our offline experiments are conducted over the real data sets collected from the \emph{Guess What You Like} of Taobao App.
Because we have stored the bid prices of the advertisers as well as the estimated indices from RTP modules for each candidate product,
we are able to train and test our algorithm by replaying the data in an offline fashion.
Similar settings can be found in other research works~\cite{cai2017real,wang2017ladder,Optimizedcostper2017Zhu,perlich2012bid,Optimalrealtime2014Zhang}.
It is worth noting that although the results of the advertising and recommendation platforms can be simulated via replaying the real data, we cannot get the exposure effect to respond to different delivered results. 
Therefore, for each target product $g$, we use its real multi-day data to fit the specific form of the exposure effect function $\mathcal{U}_g$ through a non-parametric model and then embed it in our offline flow to simulate the impact of the product's exposure on the recommended score ($pctr_{rec}$ in our settings).
Based on the log data, we collect the information of $K=186$ products with stable traffic trend.
A uniformly sampled fraction of the data are used as the training data and another fraction as test data.
All results reported are based on the test data.
In addition, we set $\eta_k=1$ for all target products and $\gamma=0.9$.

\subsection{Performance Comparison}
\begin{figure}[tb]
	\centering
	\includegraphics[width=\linewidth]{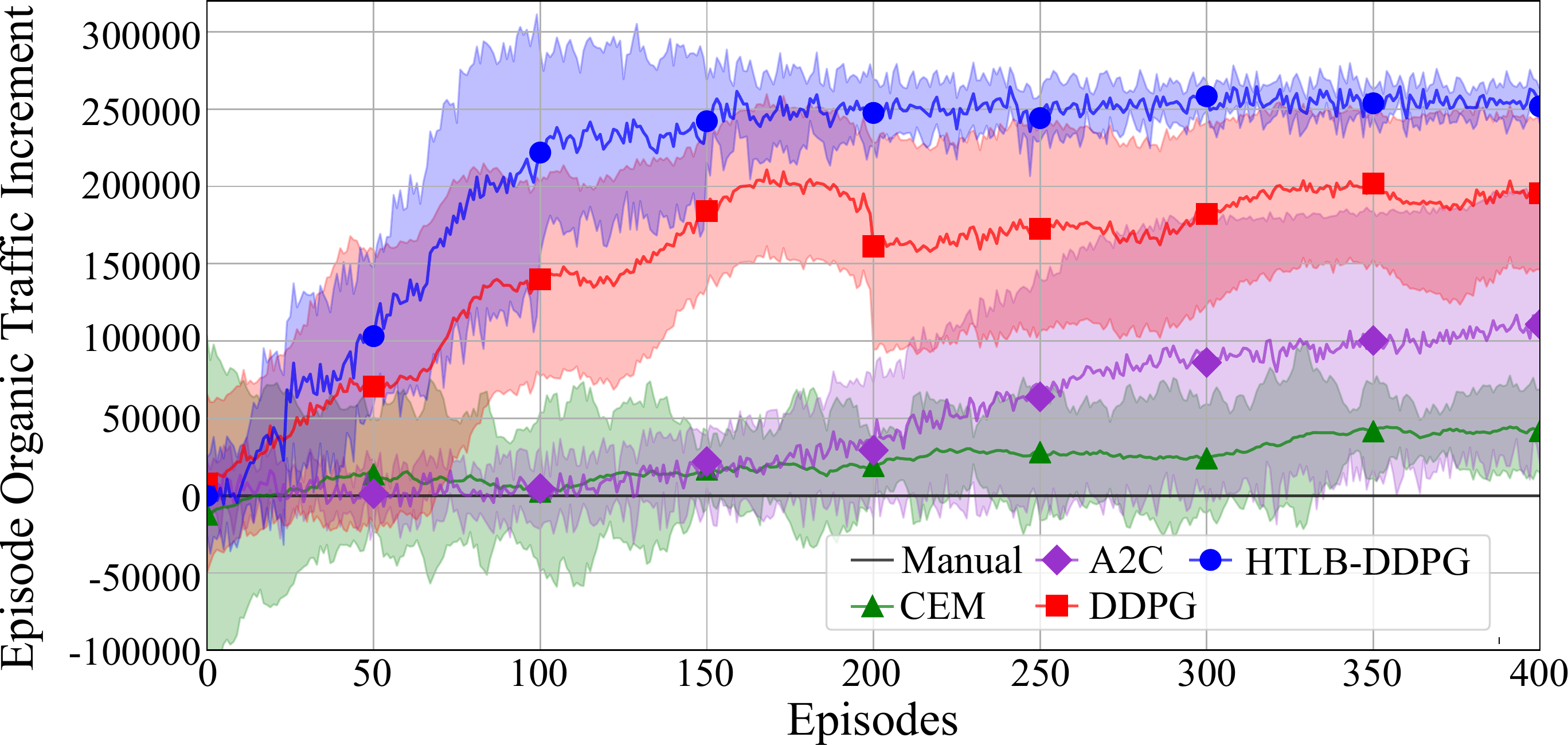}
	\caption{Learning curves compared with baselines.}\label{fig:learning_curve}
\end{figure}

With the same evaluation flow, the following algorithms are compared with our HTLB-DDPG algorithm.
We have conducted five times of experiments for each algorithm, and the average performance and standard deviations are reported in Figure~\ref{fig:learning_curve} and Table~\ref{tab:exp}.


\textbf{Manually Setting Bids}. They are the real bids set manually by the advertisers according to their experiences.
This baseline indicates the organic traffic that the advertisers can obtain with the initial fixed bid.
We use the \textbf{episode organic traffic increment compared to this baseline} as the evaluation indicator for other algorithms.
This metric indicates how much additional organic traffic (i.e., extra leveraged traffic) is obtained through the optimization algorithm.

\textbf{Cross Entropy Method (CEM)}. The algorithm~\cite{gabillon2013approximate,szita2006learning} is a gradient-free evolutionary method which searches directly in the space of policies using an optimization black box.

\textbf{Advantageous Actor-Critic (A2C)}. This method~\cite{mnih2016asynchronous,sutton1998reinforcement} is an on-policy actor-critic algorithm which applies stochastic policy gradient and does not utilize a memory replay.

\textbf{DDPG}. This is the vanilla DDPG algorithm without our HTLB training method.



The learning curves of all algorithms are depicted in the Figure~\ref{fig:learning_curve}.
Each point on the curve represents the average leveraged organic traffic
that the algorithm can achieve within an episode on the test set after training a certain number of episodes.
Among all algorithms, our HTLB-DDPG algorithm has the fastest learning speed and best convergence.
Besides, our algorithm is also superior to other baselines in terms of stability.
Compared with the stochastic policy gradient method (i.e., A2C), DDPG and our HTLB-DDPG learn faster,
which verifies the advantages of deterministic policy gradient in the continuous action space.

Table~\ref{tab:exp} lists the performance after convergence of different algorithms.
Our HTLB-DDPG outperforms all other baselines, illustrating the effectiveness of our algorithm on Leverage optimization.
The CEM algorithm does worse than all other gradient-based methods, which may be because the gradient-free method has less sample efficiency.
Besides, manually setting bids perform worst as it is a non-learning baseline.

\begin{table}[tb]
	\centering
	\caption{Converged performance compared with baselines.}\label{tab:exp}
	\small
	\begin{tabular}{cr}
		\toprule
		Method & Episode organic traffic increment \\
		\midrule
		Manual & 0 \\
		CEM & $89,512 \pm 23,413$ \\
		A2C & $206,477 \pm 5,933$\\
		DDPG &  $228,610 \pm 49,138$ \\
		HTLB-DDPG & $257,755 \pm 11,589$ \\
		\bottomrule
	\end{tabular}
\end{table}

\subsection{The Superiority of HTLB Method}
In order to better observe the superiority of our HTLB training method and its scalability to other RL algorithms,
we perform another experiment: the HTLB training method is applied to A2C, namely HTLB-A2C.
Considering that A2C is an on-policy algorithm without experience replay, we need to make two adjustments at the transition expansion stage:
(i) All the exploratory actions $a_t^{(m)}$ are sampled from the target policy $\pi$ rather than $\mu$.
(ii) At each time step, the hybrid-samples generated by the advertising emulator are directly grouped into a batch to update the network's parameters instead of being stored into memory.

\begin{figure}[tb]
	\centering
	\includegraphics[width=\linewidth]{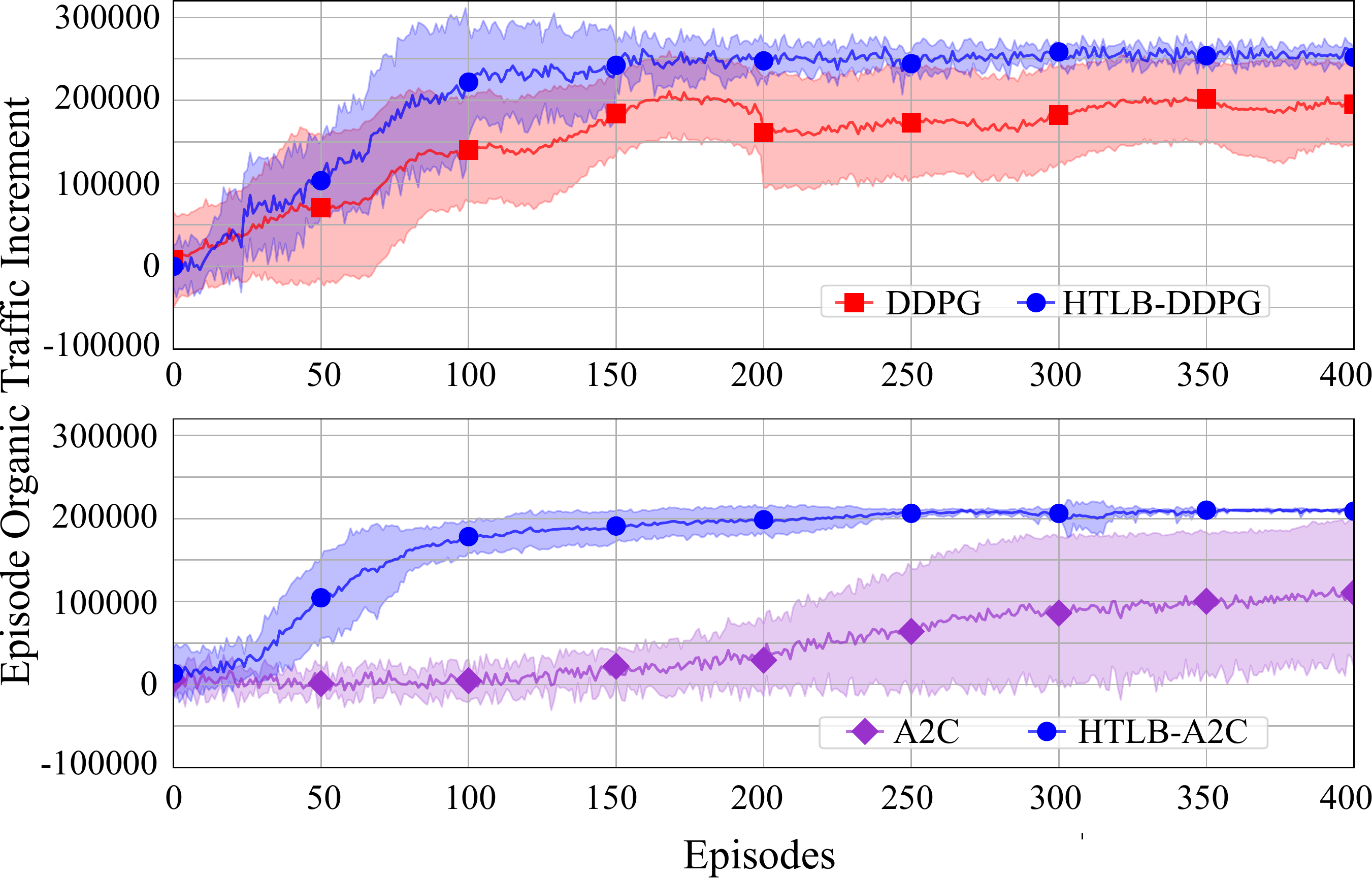}
	\caption{Learning curves of algorithms with/without HTLB.}\label{fig:htlb}
\end{figure}

As shown in Figure~\ref{fig:htlb}, we find out that the algorithms with our HTLB training method converge more stable and faster than ones without HTLB,
and there is a significant improvement on A2C,
which verifies that by introducing the HTLB training method, the sample complexity of reinforcement learning is significantly reduced.
Moreover, the HTLB training method can be applied to different RL algorithms,
and in particular, the algorithm with lower sample efficiency, e.g., A2C, can benefit more from it.

\subsection{The Impact of Business Traffic}

\begin{table}[t]
	\centering
	\caption{The impact of business traffic on organic traffic.}\label{tab:alg_analy}
	\small
	\begin{tabular}{crr}
		\toprule
		Method & Business traffic increment &Organic traffic increment \\
		\midrule
		Min bid ratio&$-448,548$&$-228,855$\\
		Manual&0&0\\
		Max bid ratio&$519,633$&$88,667$\\
		HTLB-DDPG&$196,300\pm28,185$&$257,755\pm 11,589$\\
		\bottomrule
	\end{tabular}
\end{table}

It is easy to find that optimizing bidding strategies is essentially equivalent to optimizing the business traffic allocation,
and raising the target products' bids can increase the business traffic that they can obtain when other products' bids remain fixed.
On the other hand, as analyzed in Section~\ref{sec:data-an}, we learn that most products have an increment in organic traffic when they begin to advertise.
Hence, there exists a question: \emph{can advertisers gain more leveraged traffic by merely raising bids to get more business traffic}?
To answer this question, we conduct two more experiments:
the bid adjust ratios of target products are aligned to the minimum ($-range$) and the maximum ($+range$), denoted as \emph{Min bid ratio} and \emph{Max bid ratio} respectively.
The results are reported in Table~\ref{tab:alg_analy}.

From \emph{Min bid ratio} to Manual to \emph{Max bid ratio},
the target products' bids increase gradually, and the business traffic and the organic traffic are both improved,
which verifies to some extent the positive effect of advertising on the organic traffic.
Nevertheless, the \emph{Max bid ratio} strategy is much worse than HTLB-DDPG~(88,667 vs. 257,755),
which shows that more business traffic \emph{does not imply} more organic traffic,
and our algorithm achieves effective business traffic allocation.

\subsection{Optimized Leverage Cases}
\begin{figure}[tb]
	\centering
	\includegraphics[width=\linewidth]{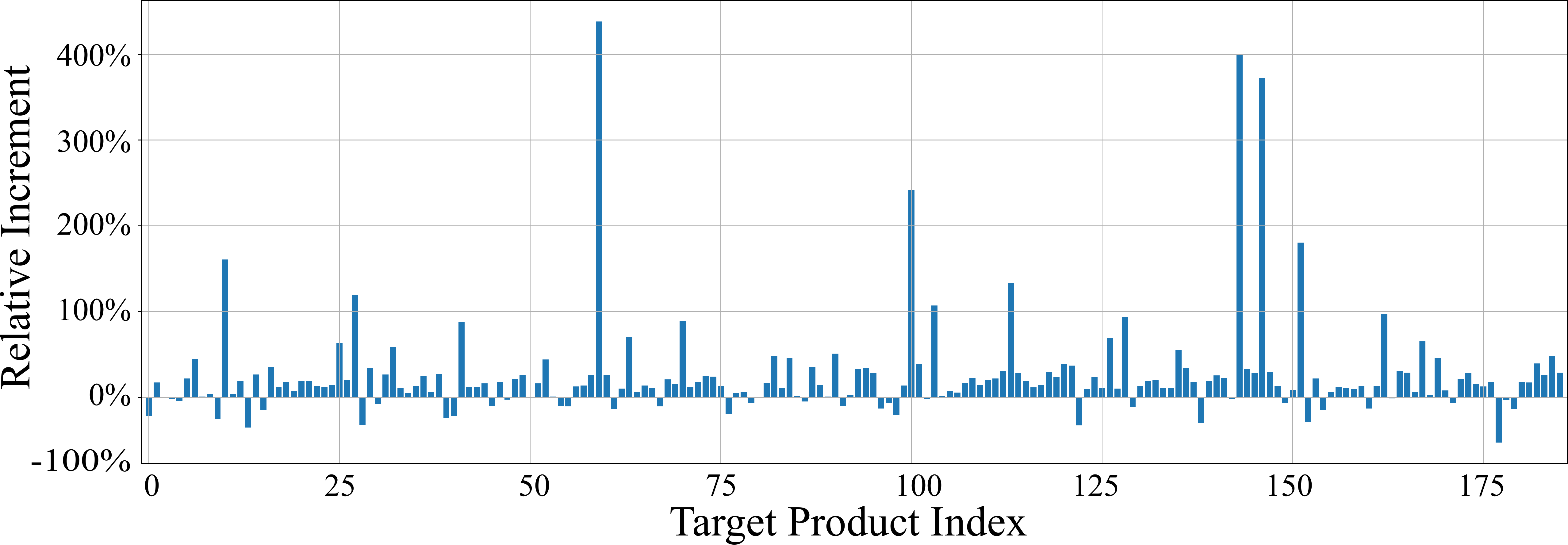}
	\caption{
		The relative increment in organic traffic obtained by each product, compared to manual setting bid.
	}
	\label{fig:all_case}
\end{figure}

In addition to studying the overall performance of target products, we have dived into the benefit that each product receives from the optimized Leverage.
Figure~\ref{fig:all_case} shows the relative increments of the organic traffic that each target product can achieve using the optimized bidding strategies compared to the fixed bidding (i.e., manually setting bid).
It is easy to find that most target products benefit from our optimization algorithm, that is, they can obtain more organic traffic than their original fixed bidding strategy.
There are even some products~(i.e., 9) that have achieved growth of more than 100\%.

\subsection{Online Results}

Next, we would like to present the results from our online deployment on ``Guess What You Like''.
We conducted an A/B test with 1,305 items which volunteered to participate in Leverage bidding.
All of the products have some Leverage potential, that is, there is still large optimization space for their organic traffic.
Considering the limitation of Taobao's online engine on large-scale gradient calculation, we use the CEM algorithm in our current online version.
Despite the use of the gradient-free algorithm, we achieve the promising results, which paves the way for the future deployment of gradient-based algorithms.

Compared to the offline version, the online algorithm has the following modifications.
(i) We use an open-source framework, Blink, to build the online stream process, which includes data collection, real-time state construction, and reward calculation.
(ii) The optimization algorithm needs to take the advertisers' budget constraints into account. Here, we clip the price adjustment ratio of each product in a rule-based way.
(iii) In order to respond to the changes of the bidding environment more quickly, the time window of decision-making is defined as one hour instead of one day.
(iv) Due to the positive correlation between the recommended score and the organic traffic, the auxiliary rewards~(i.e., the increment of recommended score) are set to reduce the difficulty of policy learning.

The 1,305 items are randomly divided into two buckets.
Bucket A is for our approach, and bucket B is for the currently deployed baseline.
Besides, we divide these items into different groups according to their organic traffic levels, as shown in Table~\ref{tab:traffic-level}.
Then we report the gap of relative increment of organic traffic between the two buckets in Table~\ref{tab:increment-gap}.
As shown in Table~\ref{tab:increment-gap}, the gap changes randomly before 20180922 but the organic traffic begins to slide toward the products in bucket A after our algorithm is online.
Three days later, the gap of relative increment between two buckets increases significantly.
The experimental results demonstrate the existence of Leverage mechanism and the control ability of our bidding algorithm on organic traffic in the online system.

\begin{table}[htbp]
	\caption{
		The number of items in different buckets with different organic traffic levels.
	}\label{tab:traffic-level}
	\small
\begin{tabular}{|l|c|c|c|c|c|c|c|c|}
\hline
traffic level   & \multicolumn{2}{c|}{Level 1}  & \multicolumn{2}{c|}{Level 2}    & \multicolumn{2}{c|}{Level 3} &  \multicolumn{2}{c|}{Level 4}  \\ \hline
organic traffic & \multicolumn{2}{c|}{500-1000} & \multicolumn{2}{c|}{1000-2000} & \multicolumn{2}{c|}{2000-3000}& \multicolumn{2}{c|}{3000-10000}\\ \hline
bucket          & A              & B            & A              & B             & A              & B      & A              & B         \\ \hline
number of items  & 202            & 185          & 174            & 172           & 106            & 110      & 200 & 156      \\ \hline
\end{tabular}
\end{table}

\begin{table}[htbp]
	\caption{
		The gap between the relative organic traffic increments of buckets A/B.
		The algorithm is online at 00:01 on 20180922.
 		The organic traffic on 20180919 is used as the initial traffic.
	}\label{tab:increment-gap}
	\small
\begin{tabular}{|l|c|c|c|c|}
\hline
\diagbox{date}{gap}{level}     & Level 1  & Level 2    & Level 3  & Level 4    \\ \hline
20180919 & 0.00\% & 0.00\%  & 0.00\% & 0.00\%  \\ 
20180920 & -0.72\% & 11.49\%  & -8.83\% & -0.16\%  \\ 
20180921 & 10.86\% & 0.81\%   & -0.94\% & -13.33\% \\ \hline
20180922 & 2.29\%  & 8.64\%   & -0.14\% & 2.34\%   \\ 
20180923 & 1.06\%  & 10.17\%  & -1.06\% & -3.99\%  \\ 
20180924 & 46.64\% & 17.79\%  & 23.76\% & 16.49\%  \\ 
20180925 & 64.71\% & 49..57\% & 48.80\% & 50.94\%  \\ \hline
\end{tabular}
\end{table}


\section{Conclusion}\label{sec:con}

The Leverage provides a new perspective on existing bidding algorithms:
by adjusting the bidding strategies, advertisers can optimize the performance of the products not only on the advertising platform but also on the recommendation platform.
In this paper, we first disclosed the Leverage mechanism in the e-commerce product feeds, revealed the principle of Leverage, and formulated the Leverage optimization problem.
Then we analyzed the characteristics of Leverage optimization and proposed an optimization algorithm that utilized Leverage to help advertisers maximize their organic traffic.
Finally, we verified our algorithm through extensive offline evaluation and deployed it to Taobao's online advertising system.
Our future works might consider optimizing more recommendation's indices such as clicks and purchases.

\bibliographystyle{ACM-Reference-Format}
\bibliography{./leverage_reference.bib}
\appendix
\section{More explanation about Leverage Phenomena}
\begin{figure}[htpb]
	\centering
	\includegraphics[width=\linewidth]{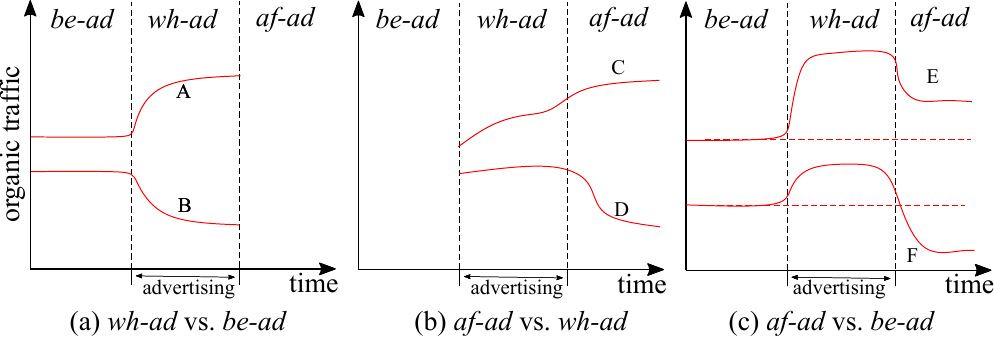}
	\caption{
		The explanatory diagram for Table~\ref{tab:lever_pheno}.
	}
	\label{fig:lever_case}
\end{figure}

In the main text, we report the number of products with different Leverage phenomena through a table.
Here we re-display it as shown in Table~\ref{tab:lever_pheno} and give more explanation for the comparison of each column in this table.
Figure~\ref{fig:lever_case} is the explanatory diagram for Table~\ref{tab:lever_pheno}.
Each curve represents a type of trend of organic traffic.
Curve A indicates an \emph{increase} in organic traffic after the delivery of advertisement, while curve B indicates a \emph{decrease}.
We divide the relative values of increase and decrease into different intervals, count the number of products in each interval, and record them in ``\emph{wh-ad} vs. \emph{be-ad}'' column of Table~\ref{tab:lever_pheno}.
Through this comparison, we can explore the impact of advertising campaigns on the organic traffic.
Similarly, curves C and D represent the trend of rising and falling, respectively, after the advertisement is withdrawn.
The two curves indicate that different items will behave differently after the advertisement is removed.
Curves E and F are only for those items whose organic traffic has a significant \emph{af-ad} drop.
Here curve E still performs better than \emph{be-ad}, and curve F is on the contrary.
Curves E and F contain 171 and 131 items, respectively
\begin{table}[hpb]
	\caption{
		The number of products with different Leverage phenomena.
		Here, 52 and 250 are bolded to indicate that a total of 302~($52 + 250$) items have a significant \emph{af-ad} organic traffic drop.
		The statistics in the last column~(i.e., \emph{af-ad} vs. \emph{be-ad}) are only for these items, i.e., $131 + 171 = 52 + 250$.
	}\label{tab:lever_pheno}
\small{
\begin{tabular}{@{}cccc@{}}
\toprule
\% improve&\emph{wh-ad} vs. \emph{be-ad} &\emph{af-ad} vs. \emph{wh-ad}&\emph{af-ad} vs. \emph{be-ad} \\
\midrule
$<-50\%$&11&\textbf{52}&\multirow{3}{*}{131}\\
$(-50\%,-10\%]$&192&\textbf{250}&\\
$(-10\%,0]$&73&97&\\\midrule
$(0,10\%]$&53&87&\multirow{3}{*}{171}\\
$(10\%,50\%]$&209&197&\\
$>50\%$&462&317&\\\bottomrule
\end{tabular}}
\end{table}

\section{Network Structure of Our Algorithm and Baselines}

Except in the manually setting bids baseline, our HTLB-DDPG algorithm and other baseline use neural networks as approximators.
The actor and critic of all algorithms include two hidden layers with 100 neurons for the first and 50 for the second.
In our implementation, the continuous action space is uniformly discretized into ten discrete actions for A2C,
so the output of the actor is the probability of each discrete action.
The activation function for hidden layers is \emph{relu}, for output layer for continuous actor is \emph{tanh}, for output layer for discrete actor is \emph{softmax} and for output layer of critic is \emph{linear}.
\section{Offline Experiment Details}
\subsection{State Setting}
As described in main text, the agent's state can be decomposed into two parts $[\mathbf{o}, \mathbf{x}]$.
For a specific product $g$,  the $t$-th time-step state contains $o_t^g$ and $x_t^g$.
The advertisement-related state $o_t^g$ contains two types of features: the description feature for product $g$ and the statistical characteristic in time window $t$.
Specifically, the description feature is to distinguish the different products,
while the statistical characteristic is used to indicate the status of the same product at different time steps.
In our settings, a product's advertising description feature is defined as a tuple $\mathcal{O}^g = (apctr_{ad}^g, apcvr_{ad}^g, bid^g, ppb^g)$,
where $apctr_{ad}^g$ (or $apcvr_{ad}^g$) is the average predicted click-through (or conversion) rate from the advertising RTP module, 
and $bid^g$ is the fixed bid price set by the advertiser and $ppb^g$ is the transaction price of the product $g$.
Such a definition for the description feature is based on two reasons:
(i) The $apctr_{ad}^g$ and $apcvr_{ad}^g$ output by the advertising RTP module are highly generalized for the product's quality.
Using the two values instead of the product's original features can transfer the knowledge learned by the existing machine learning model (i.e., the advertising RTP module) and reduce learning complexity of our problem.
(ii) Each element in $\mathcal{O}^g$ is highly correlated with the ranking rule in the recommendation and advertising platforms.
The statistical characteristics in $o^g_t$ contains the business traffic attributes obtained by $g$ in $t$-th time step,
and particularly, $pv_{t,ad}^g$, $click_{t,ad}^g$ and $ctr_{t,ad}^g$ (i.e. $click_{t, ad}^g / pv_{t, ad}^g$) in advertising platform are considered in our settings.

Different from $o_t^g$, the recommendation-related state $x_t^g$ only contains the statistical characteristics in recommendation platform (i.e., $pv_{t,rec}^g$, $click_{t,rec}^g$, $ctr_{t,rec}^g$),
because we cannot obtain the description feature of $g$ from the recommendation system.
Besides, the differences in the statistical characteristics between two adjacent time steps are also important features,
as they reflect the changing trend of the traffic obtained by the product.
Therefore, we include the features into the agent state.

\subsection{Training Parameters}
In our HTLB-DDPG algorithm, we use Adam algorithm to optimize the parameters of actor and critic.
The learning rate of actor is 0.001, while the learning rate of critic is 0.0001.
The times of expanding transition is set to 10.
The size of replay memory is 10000 and we use Ornstein–Uhlenbeck process as the random process $\mathcal{N}$ to produce exploratory actions.
The soft replacement parameter of target network $\tau$ is 0.01.
In addition, we introduce the L$_2$ regulation term to the critic loss, and the coefficient for regulation term is 0.01.

\section{Online deployment system}
\begin{figure}[htpb]
	\centering
	\includegraphics[width=0.9\linewidth]{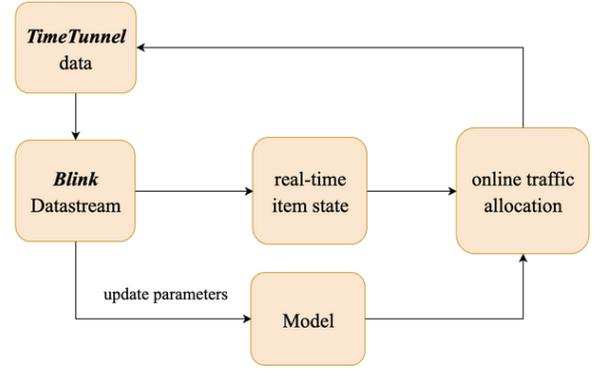}
	\caption{
		The architecture of our online deployment.
	}\label{fig:online-system}
\end{figure}

Figure~\ref{fig:online-system} shows the architecture of our online deployment.
Here, we use TimeTunnel to collect online data.
TimeTunnel is an open-source data transmission platform based on \emph{thrift} communication framework, featuring high performance, real-time, high reliability and high scalability.
Besides, Blink is utilized to build real-time online data streams.
Blink is an open-source data stream framework maintained by Alibaba on the basis of Apache Flink, and has higher performance and better stability.

\section{More Online Results}

\subsection{The proportion of products with increased organic traffic}
\begin{figure*}[htpb]
	\centering
    \hfill
	\begin{subfigure}{0.49\linewidth}
		\includegraphics[width=0.8\linewidth]{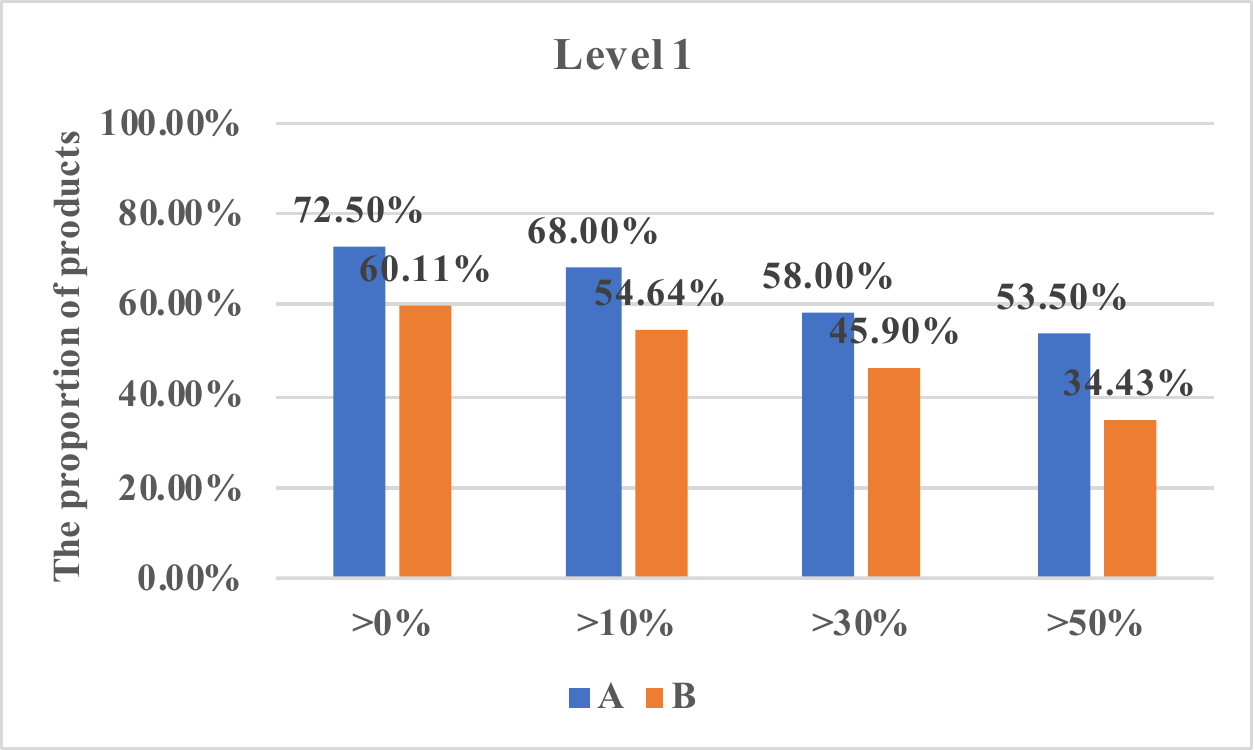}
		\caption{}\label{fig:online-ratio1}
	\end{subfigure}
	\begin{subfigure}{0.49\linewidth}
		\includegraphics[width=0.8\linewidth]{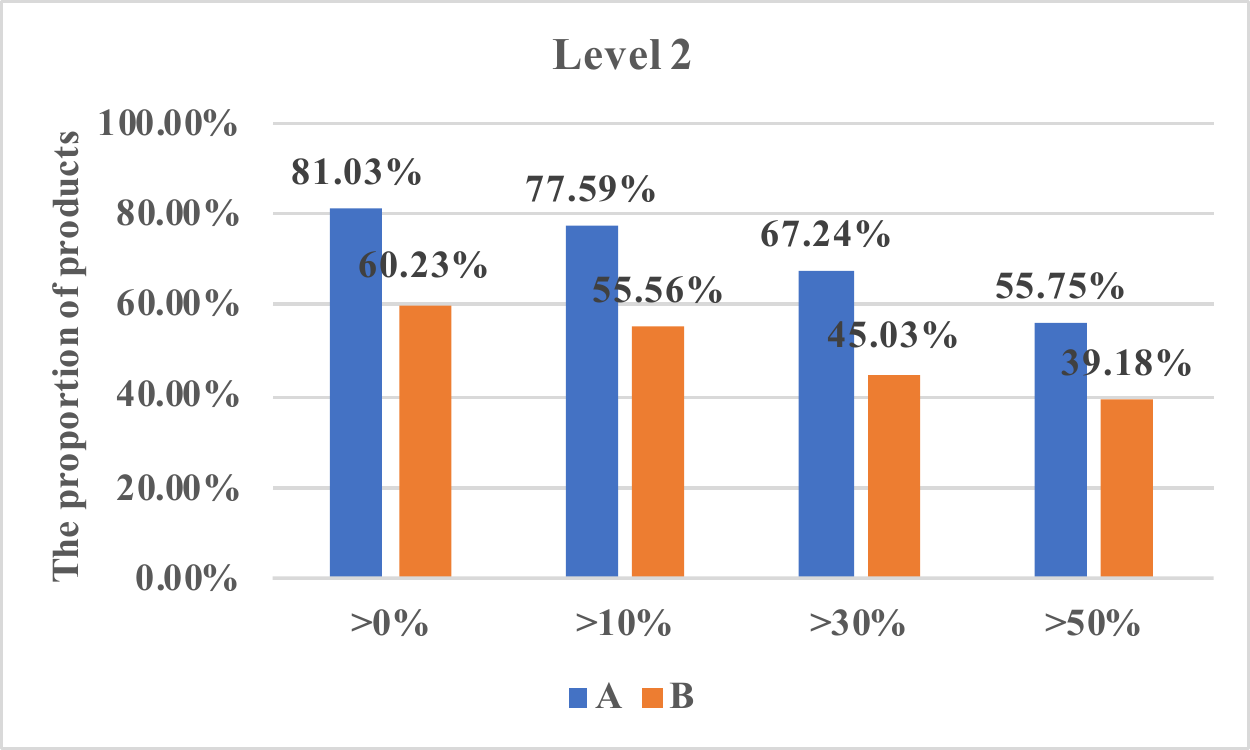}
		\caption{}\label{fig:online-ratio2}
	\end{subfigure}

    \hfill
	\begin{subfigure}{0.49\linewidth}
		\includegraphics[width=0.8\linewidth]{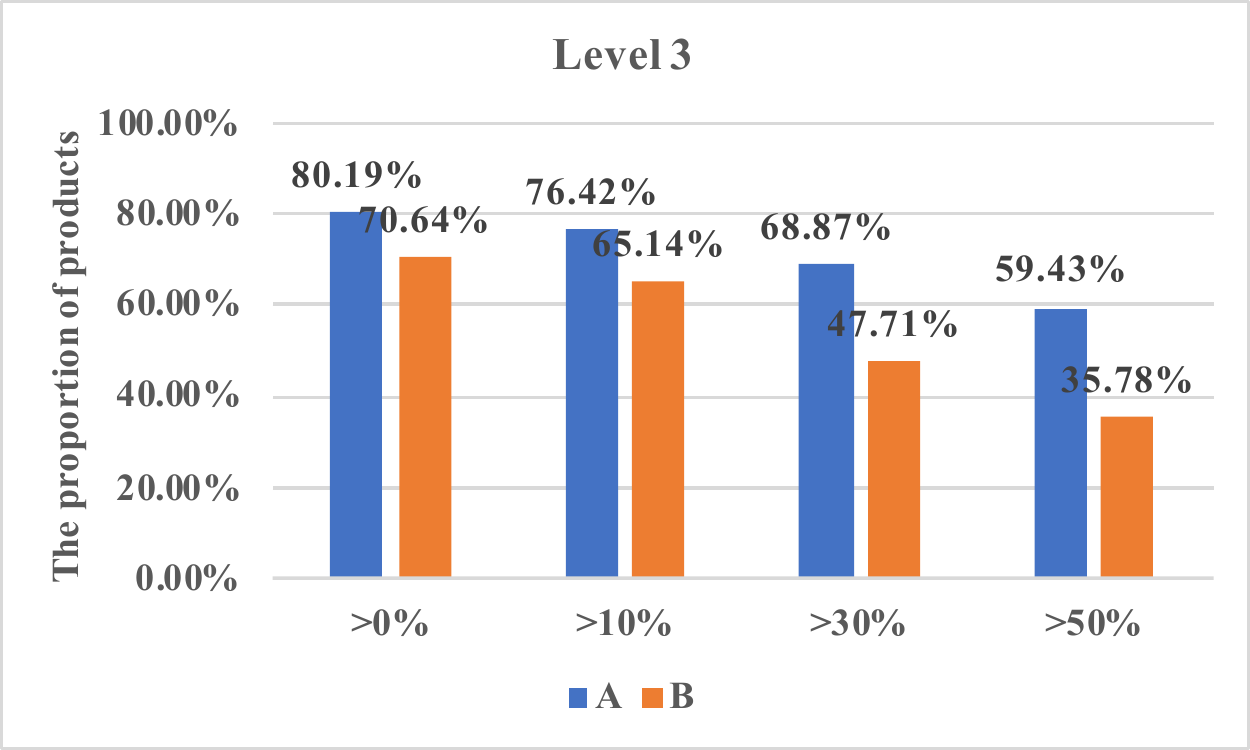}
		\caption{}\label{fig:online-ratio3}
	\end{subfigure}
	\begin{subfigure}{0.49\linewidth}
		\includegraphics[width=0.8\linewidth]{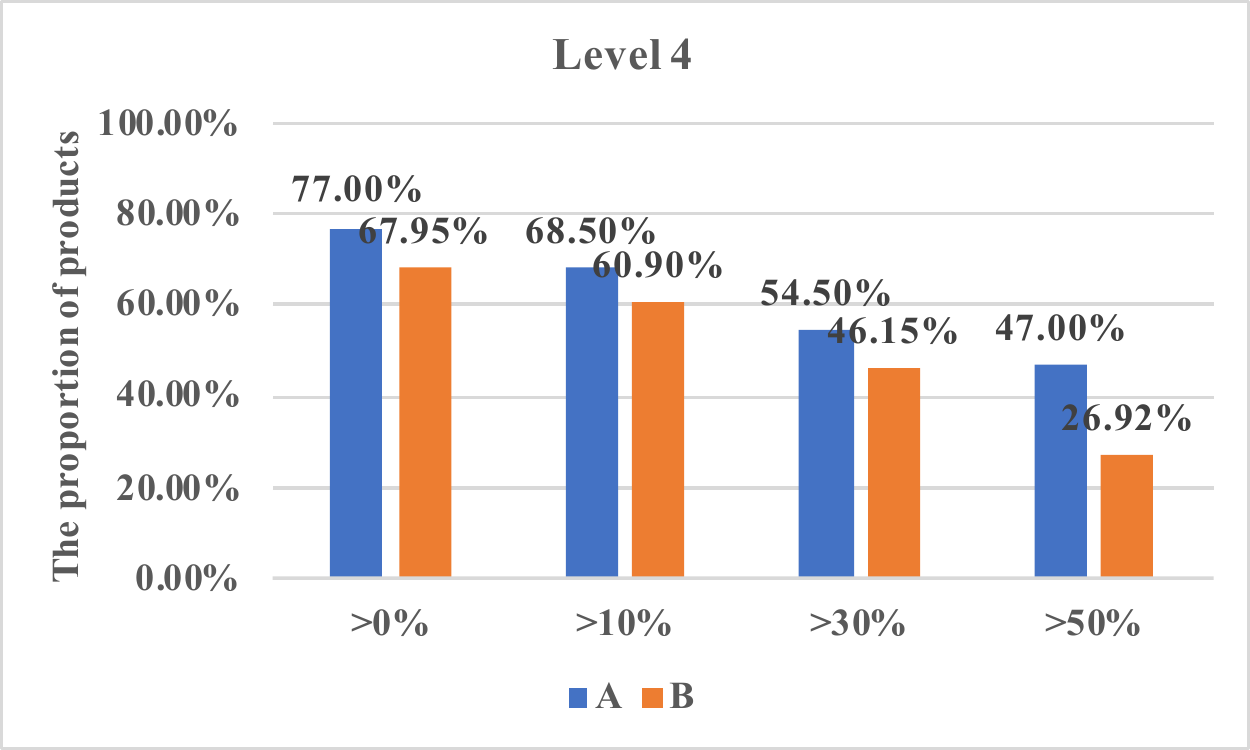}
		\caption{}\label{fig:online-ratio4}
	\end{subfigure}
	\caption{
		The proportion of products with different organic traffic changes.
	}\label{fig:proportion}
\end{figure*}

In the main text, we have compared the organic traffic between buckets A/B during 20180919$\sim$20180925.
Here we would analyze the proportion of products in each buckets with different organic traffic changes on 20180925.
The relative increment counted includes $>0\%$, $>10\%$, $>30\%$ and $>50\%$.
We compare the results of the A/B buckets in Figure~\ref{fig:proportion}.
It is easy to see that the proportion of products with increased organic traffic of bucket A is significantly more than that of bucket B, especially with a relative increment of more than 50\%.

\subsection{Online results on other dates}
This subsection will supplement the results of online experiments on other dates, i.e., 20190102$\sim$20190107.
We randomly divide 903 items into two buckets of A/B, where bucket A is for our approach and bucket B is for currently deployed baseline.
Similar to the online experiments in the main text, these items are divided into different groups according to their organic traffic levels as shown Table~\ref{tab:traffic-level}.
We observe the organic traffic changes of these products.
The organic traffic on 20190102 is recorded as the initial traffic $p^A_0$ or $p^B_0$, where $A/B$ indicates the bucket index.
The gap of relative increment of organic traffic between two buckets~(the calculation can be written as Equation~\ref{eq:gap}) is reported in Table~\ref{tab:increment-gap}.

\begin{equation}
	gap = \left(\frac{p^A_t - p^A_0}{p^A_0} \times 100\%\right) -\left( \frac{p^B_t - p^B_0}{p^B_0}\times 100\%\right)
	\label{eq:gap}
\end{equation}

\begin{table}[htbp]
	\caption{
		The number of items in A/B buckets with different organic traffic levels.
	}\label{tab:traffic-level}
	\small
\begin{tabular}{|l|c|c|c|c|c|c|}
\hline
traffic level   & \multicolumn{2}{c|}{Level 1}  & \multicolumn{2}{c|}{Level 2}    & \multicolumn{2}{c|}{Level 3}     \\ \hline
organic traffic & \multicolumn{2}{c|}{500-1000} & \multicolumn{2}{c|}{1000-2000} & \multicolumn{2}{c|}{2000-10000} \\ \hline
bucket          & A              & B            & A              & B             & A              & B              \\ \hline
number of items  & 165            & 193          & 144            & 167           & 118            & 116            \\ \hline
\end{tabular}
\end{table}
\begin{table}[htbp]
	\caption{
		The gap of the relative organic traffic increments between buckets A/B.
		Our algorithm is online at 20:00 on 20190103.
		The organic traffic on 20190102 is used as the initial traffic.
	}\label{tab:increment-gap}
	\small
\begin{tabular}{|l|c|c|c|c|}
\hline
\diagbox{date}{gap}{level}     & Level 1  & Level 2    & Level 3  & total    \\ \hline
20190102 & 0.00\%   & 0.00\%    & 0.00\%   & 0.00\%   \\ \hline
20190103 & -3.15\%  & -78.30\%  & 0.87\%   & -25.69\% \\ \hline
20190104 & 10.83\%  & 31.22\%   & 156.38\% & 81.87\%  \\ \hline
20190105 & 63.70\%  & -46.81\%  & 109.07\% & 41.81\%  \\ \hline
20190106 & 119.11\% & -148.78\% & 114.77\% & 17.28\%  \\ \hline
20190107 & 303.06\% & 140.83\%  & 87.12\%  & 136.27\% \\ \hline
\end{tabular}
\end{table}

\end{document}